\documentclass[runningheads]{llncs}
\usepackage[T1]{fontenc}
\usepackage{graphicx}
\usepackage{float}
\usepackage{newfloat}
\usepackage[skins]{tcolorbox}
\usepackage{booktabs}
\usepackage{multirow}
\usepackage{tabularx}
\usepackage{dashrule}
\usepackage{cite}

\newfloat{promptfloat}{!t}{pop}
\newtcolorbox[auto counter]{prompt}[2]{
    coltitle=black,
    label={prompt:#1},
    colback=gray!10,
    colframe=gray!20!gray,
    fonttitle=\bfseries,
    title=Prompt \thetcbcounter: #2,
    enhanced,
    fonttitle=\scriptsize, fontupper=\scriptsize, fontlower=\scriptsize, left=1mm, 
    right=1mm, 
    top=1mm, 
    bottom=1mm, 
    middle=1mm,
    arc=0pt,
    boxrule=0pt,
    borderline={0pt}{0pt}{dashed}, % Border style
    minipage boxed title*=-1.95em,
    attach boxed title to bottom center={yshift=2pt, yshift=0pt},
    boxed title style={enhanced, colback=white!55!white,
    boxrule=0pt, frame hidden}
}

\begin{document}
\title{Comparing Apples to Oranges: LLM-powered Multimodal Intention Prediction in an Object Categorization Task}
\titlerunning{Comparing Apples to Oranges}
% If the paper title is too long for the running head, you can set
% an abbreviated paper title here
%

%\author{First Author\inst{1}\orcidID{0000-1111-2222-3333} \and
%Second Author\inst{2,3}\orcidID{1111-2222-3333-4444} \and
%Third Author\inst{3}\orcidID{2222--3333-4444-5555}}
\author{Hassan Ali \and Philipp Allgeuer \and Stefan Wermter\thanks{The research was supported by Horizon Europe TERAIS, under Grant agreement number 101079338 and the DFG under the Crossmodal Learning (TRR-169) project.}}
\authorrunning{H. Ali et al.}
% First names are abbreviated in the running head.
% If there are more than two authors, 'et al.' is used.
%

\institute{Knowledge Technology, Department of Informatics, University of Hamburg\\
\email{\{hassan.ali, philipp.allgeuer, stefan.wermter\}@uni-hamburg.de}}
\maketitle              % typeset the header of the contribution
%

%\vspace*{-3.5ex}

\begin{abstract}

Human intention-based systems enable robots to perceive and interpret user actions to interact with humans and adapt to their behavior proactively. Therefore, intention prediction is pivotal in creating a natural interaction with social robots in human-designed environments. In this paper, we examine using Large Language Models (LLMs) to infer human intention in a collaborative object categorization task with a physical robot. We propose a novel multimodal approach that integrates user non-verbal cues, like hand gestures, body poses, and facial expressions, with environment states and user verbal cues to predict user intentions in a hierarchical architecture. Our evaluation of five LLMs shows the potential for reasoning about verbal and non-verbal user cues, leveraging their context-understanding and real-world knowledge to support intention prediction while collaborating on a task with a social robot.
\\[0.5ex] \textbf{Video:} \url{https://youtu.be/tBJHfAuzohI} %\vspace{-1.5ex}

% ideas for ICSR
% application for social robots and collaboration with robot
% multimodal integration of verbal and non-verbal
% ros-based application, mention this
% need to motivate the social robot aspect
% make it less about non-verbal proficiency and more on the multimodal aspect
% why the categorization scenario?
% focus on using LLM for reasoning

%\vspace{-3ex}
\end{abstract}

\section{Introduction}

With the rapid AI advancements, social robots are playing increasingly prominent roles in human societies. Unlike traditional robots, social robots are designed to facilitate a natural and intuitive interaction with humans, resembling human-human interaction rather than solely relying on explicit commands~\cite{hildt2021}. Human intention prediction is a core aspect of such seamless interaction and for joint human-robot task completion. It involves perceiving subtle cues from human actions, gestures, and speech to infer underlying goals \cite{zhangy2023}. However, it is challenging due to the complexity and context-driven nature of human behavior, where intentions are conveyed multimodally with verbal and non-verbal cues.

Typical intention-based systems require collecting a vast amount of data, which is impractical and restrictive, limiting the system to specific use cases, modalities, or setups~\cite{lubitz2023}. Recent work with LLMs in robotics has demonstrated their zero-shot abstract reasoning skills. Besides language proficiency, LLMs have exhibited notable inferential skills, broad real-world knowledge, and robust reasoning~\cite{yao_2023_reactsynergizingreasoningacting}. Also, they are performant at capturing context, making them a candidate for advancing intention prediction in Human-Robot Interaction (HRI). However, LLMs depend on textual input (user queries and prompts) to generate contextually relevant responses. Thus, it is challenging in robots for LLMs to replace human-aware models during the interaction~\cite{zhang2023}, even with the subsequent generation of multimodal LLMs, which still face difficulties processing and understanding basic visual information patterns~\cite{tong2024}. Thus, a reliable method for equipping LLMs with multimodal awareness of human cues is still missing. 

\begin{figure}[!t]
\parbox{\linewidth}{\centering\includegraphics[width=1\linewidth]{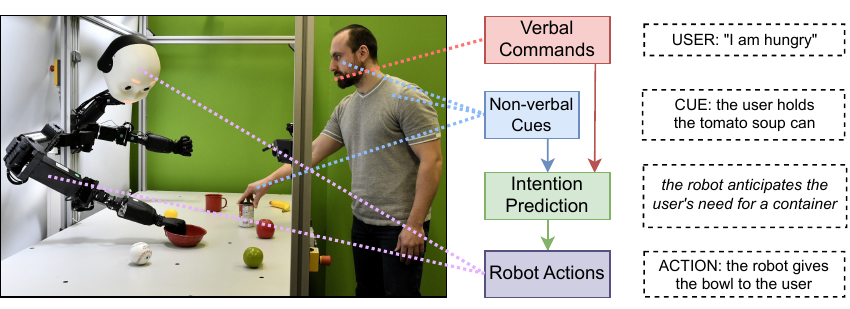}}
\caption{An overview of our intention prediction system. An LLM reasons about the user's %collected %After inferring both 
verbal (saying ``I am hungry'') and non-verbal (holding the can) cues %using an LLM, 
 to generate suitable actions complementary to those of the user, e.g., giving the bowl to the user.%\todo{replace image on the left, ex: user is pointing to the can+nicol is about to push the bowl}
}
\label{fig:setup_overview}
\vspace{-3ex}
\end{figure}

%to predict human intentions and anticipate actions
In this work, we use the Neuro-Inspired COLlaborator~\cite{kerzel2023} (NICOL) robot (see Fig.~\ref{fig:setup_overview}) to integrate multimodal user cues in an LLM-powered intent reasoning framework. We process user non-verbal cues by leveraging various machine learning vision pipelines and verbal commands via an Automatic Speech Recognition (ASR) system. In a two-layered processing method (\textit{perceptive reasoning} and \textit{task reasoning}), we use a text-based LLM to reason about the combined cues for identifying current user actions and anticipating user goals. We evaluate our system with a collaborative task for object categorization. Our experiments confirm that LLMs are viable assets for competent intention prediction. Overall, our work contributes to %both research in LLMs and robotics 
the integration of LLMs in robotics as well as human intention prediction in human-robot collaboration, effectively bridging the gap between machine understanding and the subtleties of human communication.

%\vspace{-2ex}
\section{Related Work}
\label{related_work}
%\vspace{-1ex}
Predicting user intention in social robots is challenging, given the diverse human differences and dynamic environments~\cite{lubitz2023}, and requires context comprehension and continuous user awareness~\cite{veselic2021}. Besides LLM proficiency in language, they show promising context understanding~\cite{yao2024} and adaptability to user input while possessing real-world knowledge~\cite{alkhamissi2022}. Thus, LLMs are beneficial for tasks including real-time coordination between a human and an AI agent~\cite{liu2024}, human-in-the-loop zero-shot planning~\cite{zhang2023}, belief-desire-intention driven planning~\cite{jang2023}, and human-aware motion planning in a household~\cite{graule2023}. LLMs are also useful in forecasting human behavior in video sequences~\cite{zhao2023}, and solving ambiguity in human instructions through multiple dialogue rounds for intention clarification~\cite{zhang2023_clarify}.

While clarifying intentions with conversation repair is beneficial, multimodal communication attributes to interaction intuitiveness and human understanding in AI systems. Prior work mainly used non-verbal cues for giving robots human-like abilities for apparent enhancement of the interaction. Some work included LLM grounding in social robots to generate movements and poses~\cite{yoshida2023} and improve chat agents embodiment focusing on social features like facial expressions~\cite{cherakara2023}. Other approaches used hand gestures for boosting the interaction liveliness~\cite{lee2023} or the robot's eye gaze for managing dialogue in multi-user interaction~\cite{serfaty2023}. However, there remains a gap in leveraging multimodal user cues as a natural way for humans to express their intentions. State-of-the-art approaches like~\cite{wang_lami_2024} rely on pre-defined functions and pre-programmed conditions instead of real-time contextual understanding, which limits the system's flexibility and adaptability of the interactions. In contrast, our work shows that multimodal verbal and non-verbal cues can be combined with an LLM reasoning backbone for robustly predicting human intentions, ultimately enriching the interaction.

%\vspace{-1ex}
\section{Methodology}
%\vspace{-1ex}
\begin{promptfloat}
\begin{prompt}{grounded_llm}{Our system is based on a grounded LLM as the NICOL robot~\cite{allgeuer2024}.}
\textbf{Input:}\\
%\textbf{CUE:} 
USER: Tell me about yourself.\\
%\hdashrule{\linewidth}{0.4pt}{3pt 2pt}
\textbf{Output:}\\
SAY: Hello! I am NICOL, short for "Neuro-Inspired Collaborator". I am a humanoid robot designed and built by the Knowledge Technology group of the University of Hamburg, and I am here to collaborate with you!\\
\hdashrule{\linewidth}{0.4pt}{3pt 2pt}
\textbf{Input:}\\
%\textbf{CUE:} 
USER: Give me the red object which is also a fruit.\\
%\hdashrule{\linewidth}{0.4pt}{3pt 2pt}
\textbf{Output:}\\
%\textbf{SAY:} 
SAY: I see that you're referring to the apple. Sure I can give it to you.\\
ACTION: Give apple
\end{prompt}
\vspace{-5ex}
\end{promptfloat}

%to collaborate towards a mutual goal 
%(ex: \textit{SAY} to generate speech or \textit{ACTION} to generate a robotic manipulation). 
%Our proposed architecture (see Fig.~\ref{fig:system_overview}) 
%has a dual process for intention prediction: perceptive reasoning and task reasoning. 
The system extends our previous work for LLM grounding to behave and interact as the physical NICOL robot instead of an AI language model~\cite{allgeuer2024}. This is achieved by giving the LLM facts about the surrounding environment through the robot sensors and linking the LLM's commonsense reasoning with the robot's physical actions in a modular open architecture. The framework processes user queries (user speech) and generates responses in the first person, encoding robot actions without explicit programming. The responses are parsed into tokens recognized by the robot’s ROS API platform and then mapped to concrete robot actions. For example, the token \textit{SAY} leads to generated speech, while \textit{ACTION} is translated into a robotic manipulation using the robot's arms or head (see Prompt~\ref{prompt:grounded_llm}).

\begin{figure}[!t]
\centering
\parbox{\linewidth}{\centering\includegraphics[width=1\linewidth]{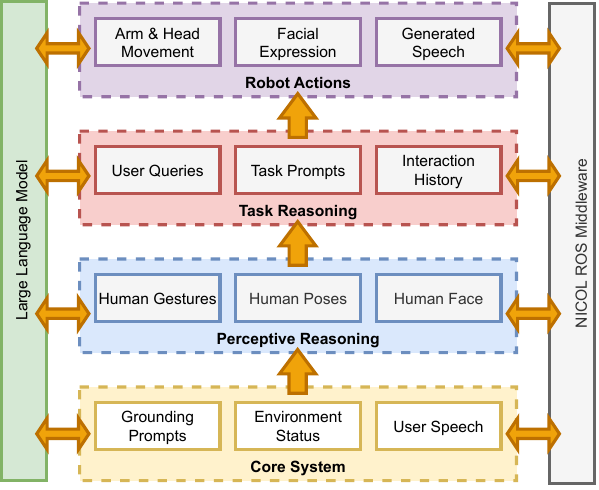}}
\caption{An overview of our system hierarchy for intention prediction. Our method for intention prediction consists of perceptive reasoning of the user's non-verbal state and task reasoning which combines explicit user queries (user speech) and task prompts.}
\label{fig:system_overview}
\vspace{-7ex}
\end{figure}

An overview of our system is shown in Fig.~\ref{fig:system_overview}. We use the core system components and robot actions as in our previous work~\cite{allgeuer2024}, where the LLM is grounded by initial prompts, the environment state is collected through the ViLD~\cite{gu2022} open-vocabulary object detector, and the user speech is recognized using Whisper~\cite{radford2022}. Also, the robot can perform a set of known actions like giving, grasping, and looking at objects as well as displaying various facial expressions. While our previous work focused on open verbal communication with an embodied agent, we contribute new modules for interpreting user non-verbal cues as well as a task reasoning framework for combining user multimodal cues to predict intentions. Our proposed method for intention prediction is a dual process consisting of perceptive reasoning and task reasoning. In perceptive reasoning, the robot collects and analyzes information about the user's non-verbal state and connects it to the proximate environment, thus formulating a set of ``beliefs'' guiding its behavior. Task reasoning incorporates explicit commands given by the user (user queries), given task prompts, and the history of the interaction in order to combine all inputs in a final LLM reasoning step for generating appropriate robot actions. The system consists of various ROS nodes, which communicate textually through designated topics with the LLM as a central component for information processing, allowing data streamlining in a standardized and uniform manner.

%\vspace{-1ex}
\subsection{NICOL Robot and Environment Overview}
We use NICOL~\cite{kerzel2023}, a semi-humanoid robot developed at the Knowledge Technology Group of the University of Hamburg. NICOL is an adult-sized robot fixed on a tabletop and has two arms and a humanoid head. Each arm has a five-fingered hand and supports bio-inspired motion planning~\cite{habekost2023} to interact with adult-sized objects. Although NICOL's design allows precise physical interaction with the environment, it also features social cues like robot facial expressions, gestures, and gaze, making it suitable for both social interaction and efficient collaboration with humans. The head features several human-like facial expressions: happiness, sadness, surprise, and anger. The NICOL API is built on top of ROS, which facilitates the integration of Python-based tools and models.

%\vspace{-1ex}

%while non-verbal cues are streamlined from the corresponding non-verbal detection and recognition modules.

%Due to LLM robust grounding to the robot's environment and collaboration setup, it provides concise textual information phrased as user facts, useful for the next reasoning step. Fig.~\ref{fig:internal_reasoning} shows the steps with concrete examples. For this step, we consider proximity gestures only.

%We combine the verbal commands, non-verbal cues, and the environment state and convert them into texts to prompt the LLM for predicting the user's intention.
\subsection{Non-verbal Cues Detection and Recognition}
%~\footnote{https://developers.google.com/mediapipe/solutions/tasks}
Human-aware robotic systems require perceiving not only user speech but also non-verbal cues~\cite{ali2023}. Thus, detecting and recognizing human gestures and body language are essential components of our approach. Due to the real-time demands of robotic systems, we utilize the open-source Tasks API of MediaPipe~(MP)~\cite{lugaresi2019}, which offers GPU-compatible computer vision models. MP was originally developed for mobile devices; however, our implementation using the NICOL robot shows its suitability for setups where the user collaborates with the robot in its proximate environment. MP constructs a pipeline of directed graphs that manage various data streams and perform specific calculations like image transformation, limb detection, limb tracking, or a machine learning inferential task using TensorFlow and Caffe, transforming the sensory input into body landmarks. We construct three MP pipelines to extract hand gestures, body poses, and facial expressions. The camera input is fed into different ROS nodes of the detection modules, extracting joint keypoints of the hand, pose, and face. Then, the user joints of each detection module are passed into a corresponding recognition module, each generating output as basic textual tokens about the non-verbal cues as expected by the LLM, e.g., the detected limb and recognized movement.

\vspace{-1ex}  
\paragraph{Hand Gestures}
 after detecting the user's hands with a single-shot Convolutional Neural Network (CNN), MP predicts 21 3D keypoints based on the GHUM model for human shapes and poses, including four for each finger and one for the wrist. We utilize these landmarks to train a gesture recognition model based on a two-layered feedforward network for the following gestures: pointing, holding an object, and resting. Pointing is with an extended arm and index finger indicating an object on the table or the table itself, holding is with the fingers curled around an object, and we define resting as the hand positioned on the side of the table. Our gesture module can robustly recognize the user's gesture using the left hand, right hand, or both. Also, the gesture detector can locate hands at various distances but we focus on the hand movements while the user collaborates with the robot using a set of objects on a table. Therefore, we consider proximate gestures that the user does with the hands over the table or within a defined physical area near the table's edge and we filter out gestures far from the table.

\vspace{-1ex}

\paragraph{Human-Pose} MP estimates user poses in a top-down way by detecting the human in the frame with an optimized MobileNetV2. User poses are estimated with 33 landmarks representing the joints with a BlazePose model. The extracted skeleton is mapped into 3D coordinates of the torso, arms, legs, and head. We utilize the 3D keypoints to provide the system with two features. First, we track the user's location in real-time relative to the robot using an average of the head keypoints. Therefore, the system can perform human-aware actions like looking at the user and lively interacting with the user based on their location in the workspace. Second, we use the pose's location to detect user proximity relative to the robot's table, i.e., whether the user is present in front of the robot.
\vspace{-1ex}

  \paragraph{Facial Expressions} 
  we utilize MP's face detector which uses a BlazeFace model based on a single shot detector to identify the face region (user's eyes, mouth, and nose) and convert it into a face mesh model. The model outputs 478 keypoints representing 3D face landmarks and a Blendshape model with 52 scores representing various facial features. We use these facial features to train a simple two-layered feedforward network for recognizing the user's neutral and smiling facial expressions, which allows the robot to react to basic user emotional states.
%\end{enumerate}

\vspace{-1ex}

\begin{figure}[!t]
\parbox{\linewidth}{\centering\includegraphics[width=1\linewidth]{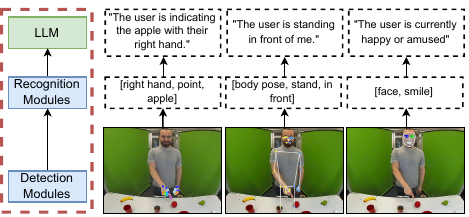}}
\caption{Perceptive reasoning of the user's non-verbal cues with examples. The user's hand, pose, and face are detected. Then, the corresponding non-verbal cues are recognized as textual tokens, passed to an LLM to generate contextually relevant outputs.%\todo{update the user images}
}
\vspace{-2ex}
\label{fig:internal_reasoning}
\vspace{-1ex}
\end{figure}

\subsection{Perceptive Reasoning}

The textual tokens generated by the non-verbal recognition modules are alone insufficient for predicting intentions since they are not grounded in the context of the interaction and setup. Since the LLM initial prompts contain information about the robot's sensory perception, surrounding environment, and possible multimodal user cues, the LLM can transform these textual tokens into coherent and contextually relevant phrases. For example, when the LLM is prompted as follows: \textit{Convert the following pose cues into a single and short meaning sentence describing the user} for the input of \textit{[bodypose, stand, in front]}, it generates output like ``the user is standing in front of me''. The LLM also has knowledge of the environment state provided as a real-time information stream of objects on the robot's table from the ViLD open-vocabulary object detector. Therefore, it comprehends the relationship between the user's gesture and objects on the table, when applicable. For example, the following cues about the user gesture: \textit{[right hand, point, apple]} are converted to ``the user is indicating the apple with their right hand''. Fig.~\ref{fig:internal_reasoning} summarizes the steps of perceptive reasoning with concrete examples. For this process, we consider proximate gestures only and the prompts are internal, invisible to the user, and meant to convert user states into meaningful textual representations for serving the next task reasoning step. 

\begin{promptfloat}
\begin{prompt}{task_prompt}{An example of a defined \textit{task prompt} for categorizing objects.}
\textbf{SYSTEM:} You and the user should collaboratively split the objects on the table into two categories. The object location on the table represents its category. There are three different possibilities (right, center, left). Category 1 represents all the objects on the right side of the table, and Category 2 represents all the objects on the left side of the table. Objects in the center do not belong to either category yet and are still uncategorized. You will not be told the exact categories explicitly but you should infer that from the user actions. The categories are based on objects' known properties like shape, purpose, material, and color. As you realize the categories that the user is creating, you should help the user decide the category an object belongs to.
\end{prompt}
\vspace{-3ex}
\end{promptfloat}

\begin{promptfloat}
\begin{prompt}{intention_1}{The robot infers the user's goal when categorizing objects.}
- the user placed the can on the right side of the table and apple on the opposite -\\
\textbf{Input:}\\
CUE: The user is pointing to the banana\\
USER: Considering the properties and usage of this object, which of category 1 or category 2 is more fitting?\\
\textbf{Output:}\\
Considering that the banana is a fruit, it would be more fitting in Category 2, which currently contains other fruits.\\
ACTION: Move yellow banana to the left side of the table\\
\hdashrule{\linewidth}{0.4pt}{3pt 2pt}
\textbf{Input:}\\
What is your inference about the user intention from the previous interactions?\\
\textbf{Output:}\\
From our previous interactions, I infer that the user intended to categorize the objects on the table into two groups: containers and fruits.
\end{prompt}
\vspace{-3ex}
\end{promptfloat}

\begin{promptfloat}
\begin{prompt}{intention_2}{The robot predicts user intention and anticipates the next action.}
\textbf{Input:}\\
CUE: The user holds the tomato soup can\\
USER: I am hungry\\
\textbf{Output}\\
ACTION: Give bowl to user\\
SAY: Would you like me to pour the tomato soup into the bowl for you?
\end{prompt}
\vspace{-5ex}
\end{promptfloat}

\begin{promptfloat}
\begin{prompt}{features_2}{Object disambiguation and intuitive interaction using pointing gestures.}
\textbf{Input:}\\
USER: Give me the yellow object\\
\textbf{Output:}\\
SAY:
Could you please clarify which object you're referring?
Is it the lemon or the banana?\\
\hdashrule{\linewidth}{0.4pt}{3pt 2pt}
\textbf{Input:}\\
USER: The user points to the banana\\
\textbf{Output:}\\
ACTION: Look at banana\\
SAY: Ah, I see! You're referring to the banana. Sure, I can give you the banana. Here you go!\\
ACTION: Give banana to user\\
\hdashrule{\linewidth}{0.4pt}{3pt 2pt}
\textbf{Input:}\\
USER: Give me another object\\
\textbf{Output:}\\
SAY: Certainly, which object would you like me to give you?\\
\hdashrule{\linewidth}{0.4pt}{3pt 2pt}
\textbf{Input:}\\
CUE: The user points to the lemon\\
\textbf{Output:}\\
SAY: Certainly, I can give you the lemon.\\
ACTION: Give lemon to user
\end{prompt}
\vspace{-4ex}
\end{promptfloat}

\begin{promptfloat}
\begin{prompt}{features_3}{The robot detects user proximity and initiates an interaction accordingly.}
\textbf{Input:}\\
CUE: The user has both hands rested on the table\\
%\hdashrule{\linewidth}{0.4pt}{3pt 2pt}
\textbf{Output:}\\
ACTION: Look at user\\
SAY: Hello there! How can I assist you today?\\
\hdashrule{\linewidth}{0.4pt}{3pt 2pt}\\
\textbf{Input:}\\
CUE: The user is pointing to the table, the user is smiling\\
USER: Hi\\
%\hdashrule{\linewidth}{0.4pt}{3pt 2pt}
\textbf{Output}\\
ACTION: Express happiness\\
SAY: Hello! Would you like me to look at or point to a specific object on the table?
\end{prompt}
\vspace{-4ex}
\end{promptfloat}

\begin{promptfloat}
\begin{prompt}{experiment_phases}{Prompts used for system evaluation in an object categorization task.}
\textbf{Phase 1 (Task Understanding Check):}\\
USER: What are the objects in category 1?\\
USER: What are the objects in category 2?\\
USER: What do objects in category 1 have in common?\\
USER: What do objects in category 2 have in common?\\
USER: Which objects are currently uncategorized?\\
%\hdashrule{\linewidth}{0.4pt}{3pt 2pt}
\textbf{Phase 2 (Object Categorization Check):}\\
\textit{the following prompt is repeated four times:}\\
CUE: The user is pointing to the \textit{object\_name}\\
USER: Considering the properties of this object, which of category 1 or category 2 is more fitting?\\
%\hdashrule{\linewidth}{0.4pt}{3pt 2pt}
\textbf{Phase 3 (Intention Understanding Check):}\\
USER: What do you infer about the user intention from the previous interactions?\\
USER: Can you explain in detail all the choices you made for categorization?
\end{prompt}
\vspace{-7ex}
\end{promptfloat}

\vspace{-2ex}
\subsection{Task Reasoning}
%goals and intentions in the categorization.
During a user-robot collaboration, the system considers the user's non-verbal states from the perceptive reasoning step and user verbal cues, collected as utterances from an ASR system based on Whisper, while acting to achieve a mutual goal. Goals can be either open-ended or defined as tasks described in the form of \textit{task prompts}, which are exchangeable and expandable in a modular way. Prompt~\ref{prompt:task_prompt} shows a task prompt example for categorizing objects, where the goal is to collaborate with the user to sort the objects on the table into two distinct groups by moving them from the center of the table to the right and left side from the robot's perspective, representing category one and category two. When reasoning about the task, the system combines the user's non-verbal states, user explicit commands, i.e., user queries, and any given task definitions to prompt the LLM. The output of this step is a combination of robot actions like arm manipulation, head manipulation, facial expressions, and generated speech. We show an example in Prompt~\ref{prompt:intention_1} where the system must infer user-intended object categories without being explicitly told. After the user places a fruit and container on each side of the table, they instruct the robot to sort a specific object. The system then combines the environment state (location of objects on the table), user non-verbal cues (pointing gesture to the banana), user verbal cue (sorting command), and the defined categorizing task, to predict the user intention (sorting distinct fruits and containers) and generate an appropriate object manipulation action (moving the banana to the left side of the table).

\vspace{-1ex}
\section{Experiments and Evaluation}
Following the proposed hierarchical architecture, the system can infer user goals and intentions in various open-ended scenarios while socially interacting with the user. For example, the robot can assist the user by giving suitable intended objects (see Prompt~\ref{prompt:intention_2}). Also, our system exhibits various additional useful features, such as establishing joint object attention between the user and robot, utilizing the LLM's language proficiency to perform conversation repairs, and identifying object ambiguities (see Prompt~\ref{prompt:features_2}), all while interacting naturally and intuitively with hand gestures. The robot can also identify user proximity, i.e., their location relative to the robot, thus initiating the interaction and playing both active and passive roles, while reacting socially to user facial expressions (cf.~Prompt~\ref{prompt:features_3}). Since evaluating open-ended interactions is challenging, we test our system through a well-defined object categorization task, as explained next.

\vspace{-1ex}
\subsection{Object Categorization Task: Results and Discussion}

\begin{figure}[!t]
\parbox{\linewidth}{\centering\includegraphics[width=1.0\textwidth]{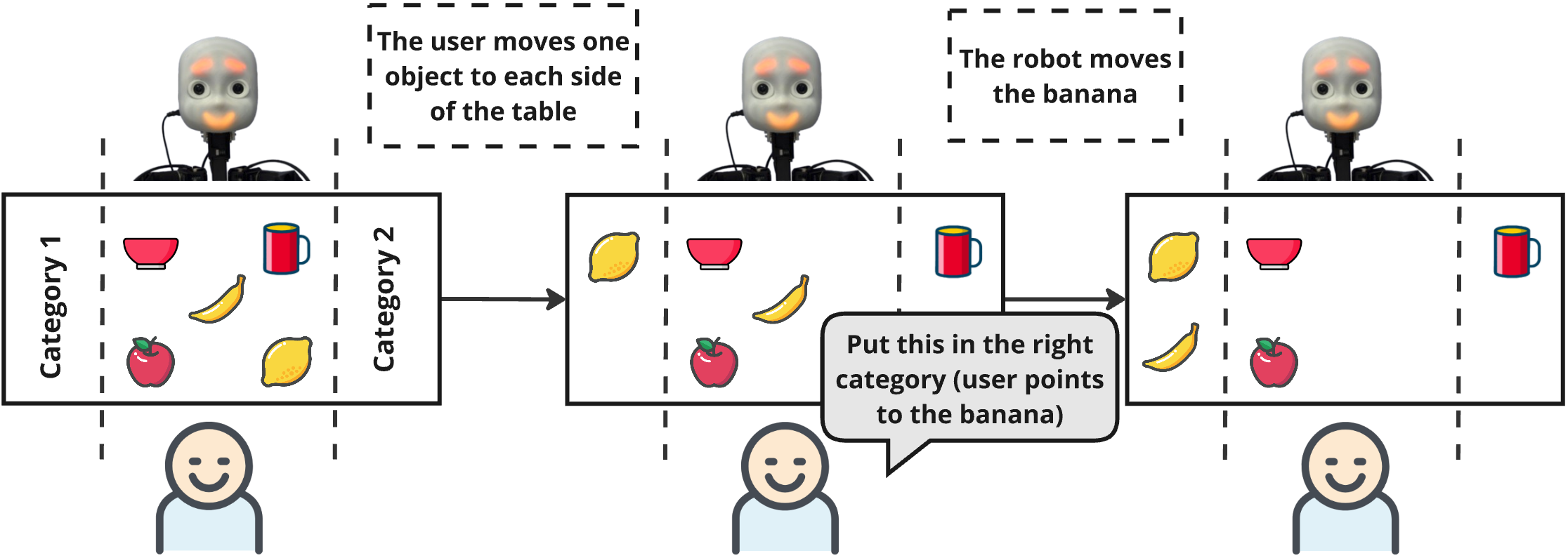}}
\caption{A concrete workflow example of the object categorization %task 
%generated 
task. After the user moves an object to each side of the table, the robot assists in categorizing the remaining objects, e.g., the banana is sorted with the lemon since both are yellow fruits.}
\label{fig:categorization_task}
\vspace{-3ex}
\end{figure}

In our task, the user and robot collaboratively categorize objects on the table into two distinct groups. The robot assists the user in finding the fit for each ``uncategorized'' object based on common properties and characteristics, like shape, color, and purpose. Unlike typical autonomous object categorization tasks in which categories are known~\cite{koskinopoulou2021, pan2021, ramadurai2022}, the robot is not given an explicit definition of the categories but should infer it based on user actions. The task starts as the user places one object on each side of the table, i.e., representing \textit{Category~1} and \textit{Category~2}. Then, the robot responds to user queries about where to place a specific uncategorized object. An example of the categorization task is in Fig.~\ref{fig:categorization_task}.

We evaluate the task with six objects: red apple, yellow banana, red can, yellow lemon, red bowl, and red cup. We select the objects due to their various purposes (fruits and containers), colors (red and yellow), and shapes (round and cylindrical). For performance comparison, we start the task with every possible object pair as category initiators. After category initialization, the robot is asked to sort a randomly selected object. We conduct 150 trials, containing every possible object pair, each with a reset chat history. Moreover, we evaluate the system using OpenAI and open-weight LLMs: GPT-3.5-turbo-0301, GPT-3.5-turbo-16k-0613, GPT-4-0613, Vicuna-1.3-33b, and Mistral-7b-instruct-v0.2. We follow a three-phase evaluation scheme (see Prompt~\ref{prompt:experiment_phases}): 1) Task Understanding Check: five prompts testing the LLM's task understanding and categories at hand, 2) Object Categorization Check: for each ``uncategorized'' object, a prompt is given as a categorization request (four prompts in total), and 3) Intention Understanding Check, in which the LLM's understanding of user intentions and goals is tested with two prompts. The LLM is also expected to logically explain its decisions and actions. A summary of the results is in Table~\ref{table:categorization_task} with values representing model accuracy, averaged by the number of trials and object pairs.

Our system exhibited proficiency in predicting user goals and deriving intended categories of the various fruits and containers, regardless of the comparison order, especially GPT-4, with highly consistent accuracy. The performance of the GPT-3.5 models was comparable and dropped slightly when comparing objects with similar functions (fruit vs. fruit or container vs. container), indicating these models gave more importance to ``usage'' over other attributes. Nonetheless, GPT-4 showed competence and flexibility with category criteria that relied on subtle features like color or shape. All OpenAI models showed good task realization in Phase 1 and the occasional accuracy drop is explained by accidentally mixing the category location (swapping right and left), hinting that the LLM spatial reasoning and orientation have not yet progressed to a performant level.

%\todo{maybe some information about task complexity and other papers that addressed it}
\begin{table}[!t]
\caption{Results of the object categorization task %using different LLM backbones.
}
\begin{tabular}{@{}cp{2.5cm}p{2cm}p{2cm}p{2cm}p{2cm}@{}}
\toprule
                                                     & \textbf{Model}     & \textbf{\begin{tabular}[c]{@{}l@{}}Fruit \\ Container\end{tabular}} & \textbf{\begin{tabular}[c]{@{}l@{}}Fruit \\ Fruit\end{tabular}} & \textbf{\begin{tabular}[c]{@{}l@{}}Container \\ Fruit\end{tabular}} & \textbf{\begin{tabular}[c]{@{}l@{}}Container \\ Container\end{tabular}} \\ \midrule
\multirow{2}{1.5cm}{\multirow{4}{*}{\textbf{Phase 1}}}   & GPT-3.5    & \textbf{0.99}                     & 0.68                 & \textbf{0.91}                     & 0.88                         \\
                                                     & GPT-3.5-16k & 0.62                     & 0.74                 & 0.89                     & 0.81                         \\
                                                     & GPT-4 & 0.95                     & \textbf{0.84}                 & 0.9                     & \textbf{0.98}                         \\
                                                     & Vicuna & 0.33                     & 0.33                 & 0.42                     & 0.39                         \\
                                                     & Mistral & 0.32                     & 0.18                & 0.21                     & 0.33                         \\[0.5ex]\hline\addlinespace[0.8ex]
\multirow{2}{1.5cm}{\multirow{4}{*}{\textbf{Phase 2}}} & GPT-3.5    & 0.82                     & 0.52                 & 0.8                      & 0.61                         \\
                                                     & GPT-3.5-16k & 0.7                     & 0.73                 & \textbf{0.98}                     & 0.78                         \\
                                                     & GPT-4 & \textbf{0.96}                     & \textbf{0.93}                & 0.97                     & \textbf{0.99}                         \\
                                                     & Vicuna & 0.54                     & 0.48                 & 0.73                     & 0.53                         \\
                                                     & Mistral & 0.725                     & 0.78                 & 0.35                     & 0.66                         \\[0.5ex]\hline\addlinespace[0.8ex]
\multirow{2}{1.5cm}{\multirow{4}{*}{\textbf{Phase 3}}}   & GPT-3.5    & 0.71                     & 0.37                 & 0.71                     & 0.7                          \\
                                                     & GPT-3.5-16k & 0.75                     & 0.53                 & 0.89                     & 0.63                         \\ 
                                                     & GPT-4 & \textbf{0.92}                     & \textbf{0.75}                 & \textbf{0.92}                     & \textbf{0.93}                         \\
                                                     & Vicuna & 0.17                     & 0.56                 & 0.48                     & 0.29                         \\
                                                     & Mistral & 0.75                     & 0.53                 & 0.44                     & 0.4                         \\
                                                     %\cmidrule(l){1-6}
                                                     \bottomrule
                                                     
\end{tabular}
\label{table:categorization_task}
\vspace{0ex}
\end{table}

\begin{table}[!t]
\centering
\caption{Pearson correlation of categorization phases (with \textit{p-values}) %using different LLM backbones.
}
\begin{tabular}{@{}p{2.5cm}p{3cm}p{3cm}p{3cm}@{}}
\toprule
\textbf{Model}       & \textbf{Phase 1--Phase 2} & \textbf{Phase 2--Phase 3} & \textbf{Phase 1--Phase 3} \\ \midrule
GPT-3.5     & 0.8 (\textbf{0.0003})     & 0.6 (\textbf{0.018})      & 0.53 (\textbf{0.044})     \\
GPT-3.5-16k & 0.53 (\textbf{0.04})      & 0.51 (\textbf{0.05})      & 0.4 (0.14)      \\
GPT-4       & 0.44 (0.1)       & 0.79 (\textbf{0.0005})    & 0.63 (\textbf{0.012})     \\
Vicuna      & 0.53 (\textbf{0.043})       & 0.75 (\textbf{0.001})    & 0.49 (0.062)     \\
Mistral     & 0.56 (\textbf{0.029})       & 0.21 (0.44)    & 0.26 (0.35)     \\ 
\bottomrule
\end{tabular}
\label{table:categorization_task_2}
\vspace{-4ex}
\end{table}

\begin{figure}[!t]
\parbox{\linewidth}{\centering\includegraphics[width=1\linewidth]{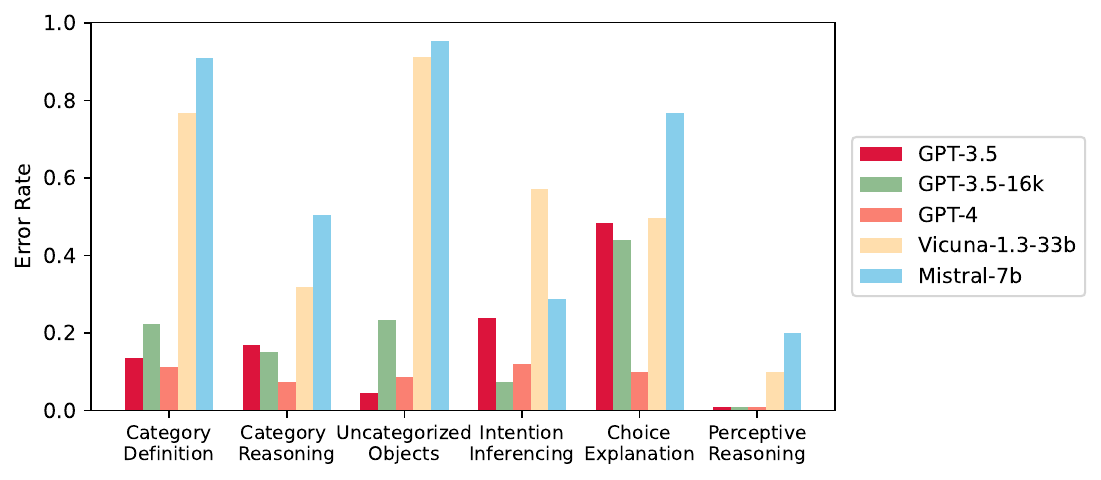}}
\vspace{-3ex}
\caption{The main sources of system errors in the object categorization task. All LLM models showed good performance in perceptive reasoning. GPT-4 showed the highest performance overall, especially in explaining the decisions made during the interaction.%\todo{update the user images}
}
\label{fig:error_rates}
\vspace{-4ex}
\end{figure}

The open-weight models came short in Phase 1 and had near-average accuracy in Phases 2 and 3. Mistral showed higher competence in Phase 2 than Vicuna despite the smaller size (7B vs. 33B parameters), which we hypothesize to be a result of its own training, i.e., Mistral is fine-tuned on instructions data making it more flexible with user commands, while Vicuna is fine-tuned on conversation data. However, our results showed a positive correlation between the model's task understanding in Phase 1 and its performance in subsequent phases (cf. Table~\ref{table:categorization_task_2}). This becomes clearer by observing the error rates (see Fig.~\ref{fig:error_rates}), which are lower for \textit{Category Reasoning} than grounding factors (\textit{Category Definition} and \textit{Uncategorized Objects}). Category Reasoning is the LLM's ability to correctly identify what a collection of objects in a certain category have in common, while Category Definition and Uncategorized Objects represent errors in the LLM's understanding of the categorization task (evaluation prompts of phase 1). 

From the error rates, we observe that the smaller models like Vicuna and Mistral are more challenging to ground to a task definition, which negatively influences their performance. However, all LLMs showed high success rates at interpreting non-verbal cues through our perceptive reasoning with marginal error rates for Mistral and Vicuna (see \textit{Perceptive Reasoning} in Fig.~\ref{fig:error_rates}). From the OpenAI models, GPT-4 reached the best performance at Phase 3, while the other OpenAI models, despite making correct decisions, were less precise at explaining decisions (see \textit{Choice Explanation} in Fig.~\ref{fig:error_rates}). Although ChatGPT models showed better resilience in task grounding, the results showcased that the highest error rates were related to not fully grasping some task-related concepts rather than the categorization itself, suggesting that the model's reasoning aligns with the task objectives. Overall, our experiment showed the system's ability to perform the assigned task, while the influence of grounding errors in smaller LLMs is expected to be mitigated as more advanced models become widely available.

%As the model develops a better task understanding over time, its performance is expected to improve. 

%We proposed a hierarchical process for intention prediction with our perceptive and task reasoning modules.
\vspace{-1ex}
\section{Conclusion}
%\vspace{-1ex}

We explored LLMs for predicting human intention, showing their ability to infer user goals and generate discrete actions to fulfill them in a human-robot collaborative task. We introduced a novel hierarchical dual process for predicting user intentions and anticipating goals by leveraging user verbal and non-verbal cues. Our perceptive reasoning module showed the high capacity of LLMs to multimodal user social cues. Through our task reasoning, the system was also able to adapt to an object categorization task, in a few-shot manner. Task prompts can be easily replaced or extended, highlighting the system's potential for rapid task acquisition and deployment in diverse settings, thus laying the groundwork for effective intention prediction in robotic collaborative and social scenarios. Future work will assess the LLM scalability across a broader range of interactive scenarios to ascertain the transferability of its intention prediction skills across multiple tasks. We will also examine the applicability of multimodal LLMs for inferring user gestures and poses in real-time social interaction with the robot.

%our perceptive and task reasoning modules, 

%We showed the high capacity of LLMs to multimodal user social cues. 

%We explored LLMs for predicting human intention, showing their ability to infer user goals and generate discrete actions to fulfill them in a human-robot collaborative task. Our approach introduced a novel hierarchical process for prediction user intentions with our perceptive and task reasoning modules, leveraging user verbal and non-verbal cues. We showed the high capacity of LLMs to multimodal user social cues. 

%Our system was also able to adapt to an object categorization task, in a few-shot manner, highlighting its potential for rapid task acquisition and deployment in diverse settings, thus laying the groundwork for effective intention prediction in robotic collaborative and social scenarios. Future work will assess the LLM scalability across a broader range of interactive scenarios to ascertain the transferability of its intention prediction skills across multiple tasks. We will also examine the applicability of multimodal LLMs for inferring user gestures and poses in real-time social interaction with the robot.

%
% ---- Bibliography ----
%
% BibTeX users should specify bibliography style 'splncs04'.
% References will then be sorted and formatted in the correct style.
%
\vspace{-2ex}
\bibliographystyle{splncs04}
\bibliography{main}

\end{document}